\documentclass[11pt]{article}

\usepackage{microtype} 
\usepackage{booktabs}  
\usepackage{url}  
\usepackage{csquotes}

\usepackage{amsmath}
\usepackage{amsthm}

\usepackage{paralist}

\usepackage[final,workshop]{automl}
\usepackage{natbib}
\usepackage{filecontents}

\usepackage{todonotes}
\newcommand{\note}[1]{\todo[inline]{#1}}
\renewcommand{\note}[1]{}  

\newcommand{\rs}[0]{\texttt{Random Search}}
\newcommand{\pb}[0]{\texttt{PriorBand}}
\newcommand{\hb}[0]{\texttt{HyperBand}}
\newcommand{\autorank}[0]{\texttt{Autorank}}
\newcommand{\lmems}[0]{\texttt{LMEMs}}
\renewcommand{\vec}[1]{\mathbf{#1}}
\newcommand{\mat}[1]{\mathbf{#1}}

\title{LMEMs for post-hoc analysis of HPO Benchmarking}

\author[1,$\ast$]{\nameemail{Anton Geburek}{gebureka@cs.uni-freiburg.de}}
\author[1,$\ast$]{\nameemail{Neeratyoy Mallik}{mallik@cs.uni-freiburg.de}}
\author[1,$\ast$]{\nameemail{Danny Stoll}{stolld@cs.uni-freiburg.de}}
\author[2]{\nameemail{Xavier Bouthillier}{xavier.bouthillier@umontreal.ca}}
\author[1,3]{\nameemail{Frank Hutter}{fh@cs.uni-freiburg.de}}

\affil[1]{Machine Learning Lab, University of Freiburg}
\affil[2]{Mila, Montreal}
\affil[3]{ELLIS Institute Tübingen}

\affil[$\ast$]{\textit{Correspondence}: \{gebureka,mallik,stolld\}@cs.uni-freiburg.de}

\hypersetup{%
  pdfauthor={AutoML}, 
  pdftitle={Documentation for the automl Package},
  pdfsubject={Documentation for automl Package},
  pdfkeywords={AutoML, LaTeX, style}
}

\begin{document}

\maketitle

\begin{abstract}
The importance of tuning hyperparameters in Machine Learning (ML) and Deep Learning
(DL) is established through empirical research and applications, evident from the increase
in new hyperparameter optimization (HPO) algorithms and benchmarks steadily added by
the community. 
However, current benchmarking practices using averaged performance across many datasets may obscure key differences between HPO methods, especially for pairwise comparisons.
In this work, we apply Linear Mixed-Effect Models-based (LMEMs) significance testing for post-hoc analysis of HPO benchmarking runs.
LMEMs allow flexible and expressive modeling on the entire experiment data, including information such as benchmark meta-features, offering deeper insights than current analysis practices.
We demonstrate this through a case study on the PriorBand paper's experiment data to find insights not reported in the original work.
\end{abstract}


\section{Introduction}\label{sec:introduction}

Hyperparameter Optimization (HPO) research has seen a surge in new benchmark contributions in recent times~\citep{eggensperger-neuripsdbt21a,pineda-neuripsdbt21a,wang-arxiv21a,pfisterer-automl22a} that have led to improved HPO algorithm contributions too.
This is a genuine attempt at making hyperparameter optimization (HPO) research an empirical and reproducible science, which is essential for the adoption of HPO in practice. 
The plethora of benchmarks can lead to large experimental data collected.
The usual modus operandi is to use relative ranks per run instance to average the results across benchmarks for a seed, with the variance of this mean across seeds accounting for the uncertainty in relative ranks, thus compressing the experiment data into one easy-to-parse aggregated plot~\citep{mallik-neurips23a}.
However, it is often observed that different problem instances can be solved best by different types of HPO algorithms~\citep{eggensperger-neuripsdbt21a}, akin to \textit{No free lunch}~\citep{wolpert-tr95a}.

In this work, we explore the application of model-based significance analysis to exploit the rich HPO experimental data from benchmarking runs.
We believe that the relative performance of different HPO algorithms are strongly hierarchical in nature when considering different benchmarks or different HPO budget horizons.
Moreover, we believe that finding sub-groups of benchmarks that capture different trends in relative performances can provide added insights.

The overall contribution of our work include:
\begin{inparaenum}
    \item Demonstrating application of LMEM-based significance analysis in HPO Benchmarking.
    \item Preset LMEM-based model recipes for sanity checking experimental data and \autorank-like analysis.
    \item Accounting for multiple benchmark metafeatures in pairwise comparison of HPO algorithms.
\end{inparaenum}
\\
Our code: \url{https://github.com/automl/lmem-significance}.

\section{Related Work and Background}\label{sec:background}

\begin{figure}
    \centering
    \includegraphics[width=0.65\linewidth]{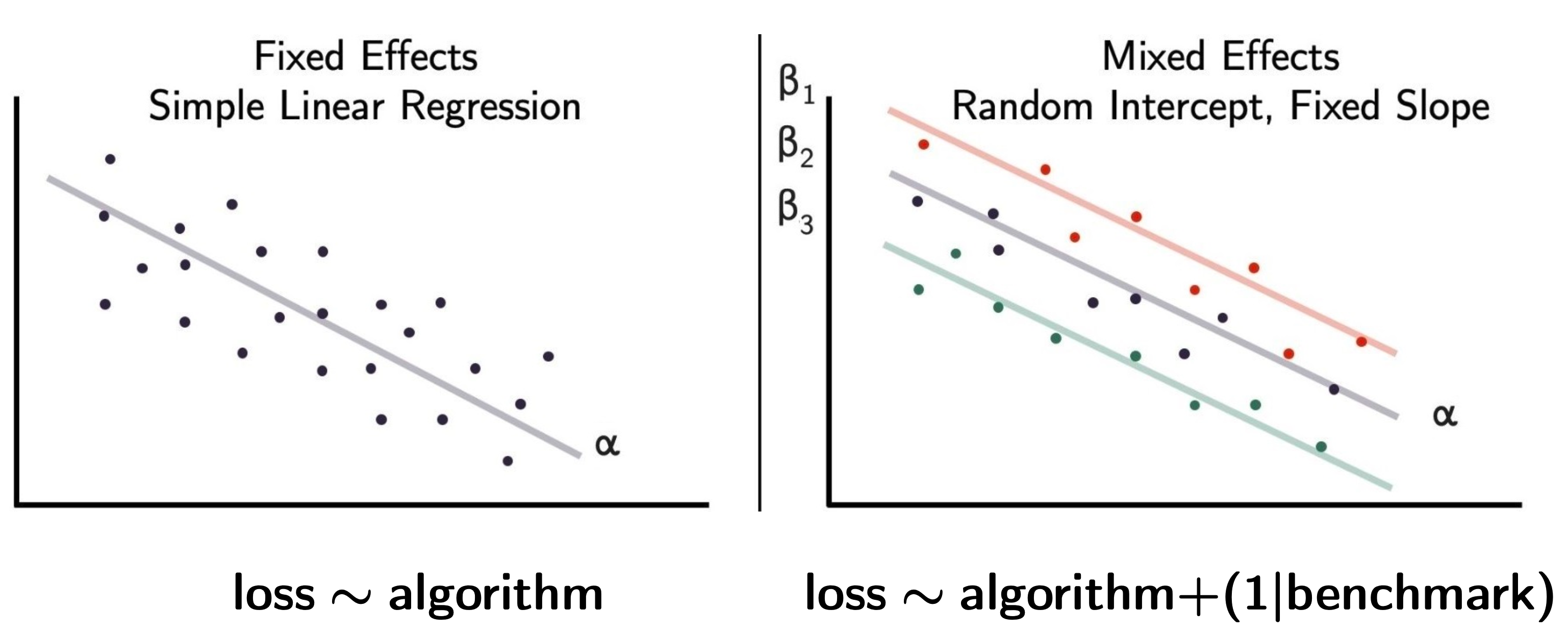}
    \caption{
    (\textit{left}) Fixed effects are fully observed and typically noise-free, i.e., loss (y-axis) recorded against algorithms (x-axis); 
    (\textit{right}) Random effects assume samples to be from some random distribution within each specific group, as described by $(1|\texttt{benchmark})$ and the $3$ lines representing $3$ groups benchmarks.
    Image sourced under \texttt{CC BY-SA 4.0}.
    }
    \label{fig:lmems-viz}
\end{figure}

LMEM-based significance testing has been brought to the attention of the Natural Language Processing community by \citep{riezler-22a} to enable the joint analysis of both different data sets and meta-parameter settings. 
Such tests were recommended especially when there was a hierarchical structure within the data.
LMEMs can account for both \textit{fixed effects} and \textit{random effects} (see, Figure~\ref{fig:lmems-viz}).
For HPO Benchmarking data, let $M_0: \texttt{loss}\sim\texttt{algorithm}$\ and $M_1: \texttt{loss}\sim\texttt{algorithm}+(1|\texttt{benchmark})$\footnote{here, \texttt{loss} is the target, \texttt{algorithm} is the fixed effect and \texttt{benchmark} is the random effect}, be two LMEM models fit on the same data.
The Generalized Likelihood Ratio test (GLRT) compares the likelihood of the data given the model, to determine if \texttt{algorithm} can significantly effect the loss after accounting for random effects from different benchmark groups.
This is in contrast to the commonly used Friedman test that doesn’t assume normally distributed data groups or homogeneous variances~\citep{demsar-06a} but also cannot handle hierarchical relations in the data. 
The Wilcoxon test is also used to compare two algorithms over multiple benchmarks, collecting each significant win, tie, or loss as a ratio~\citep{eggensperger-neuripsdbt21a}. 
\citet{dror-tacl17a} proposes to use a partial conjunction hypotheses to account for comparisons across multiple datasets, answering the question ``does on algorithm $A$ 
significantly outperform another algorithm $B$ on at least $u$ datasets out of $N$''.

For methodological details, refer to \citet{riezler-22a} and Appendix~\ref{app:lmem-glrt}.

\section{Empirical setup}\label{sec:setup}

We directly demonstrate the application and utility of LMEM-based testing through a case study on real HPO Benchmarking data.
We use experiment data from~\citet{mallik-neurips23a} containing runs of more than $5$ different HPO algorithms, on more than $30$ benchmark instances, repeated for $50$ different seeds.
For a focused study, we look at the main hypothesis and result in~\citet{mallik-neurips23a}, that shows \pb{} to be better than \hb{} under different expert prior quality scenarios (see, Figure~\ref{fig:PB_rel_rank}).
Subsequently, our focus would be on the comparison of \rs{} (RS), \hb{} (HB) and \pb{} (PB) over the benchmark instances comprising of \textit{good} (at25) and \textit{bad} expert prior input.
To serve as a simple baseline simulating current standard practice, we use \autorank{}\footnote{\url{https://sherbold.github.io/autorank/}} as the baseline to perform Friedman test on the same experiment data.
We use Critical-Difference (CD) (see, Fig.~\ref{fig:lmems-autorank}) plots to show pairwise significance difference in performance.

\section{Application}\label{sec:recipes}

In this section, we apply the method discussed to experiment data from ~\citet{mallik-neurips23a}.

\subsection{Drop-in replacement for \autorank{}}\label{sec:drop-autorank}

\begin{figure}
    \centering
    \begin{tabular}{cc}
       \includegraphics[width=0.28\columnwidth]{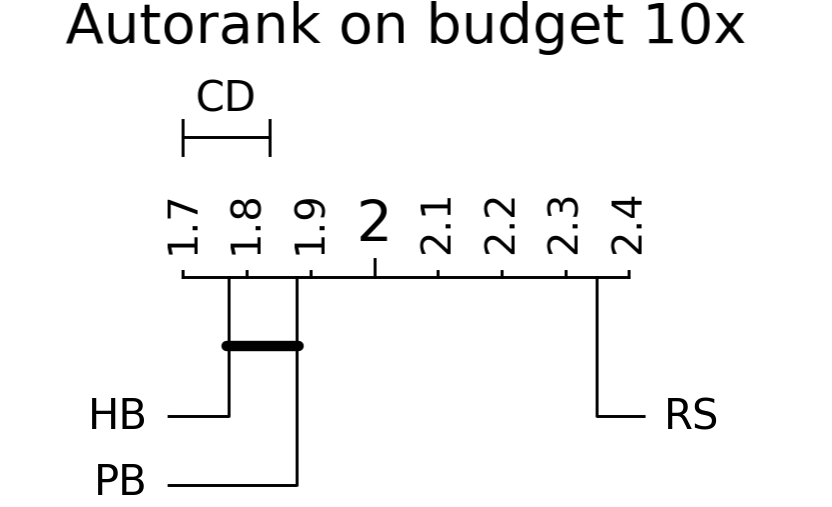} & \includegraphics[width=0.28\columnwidth]{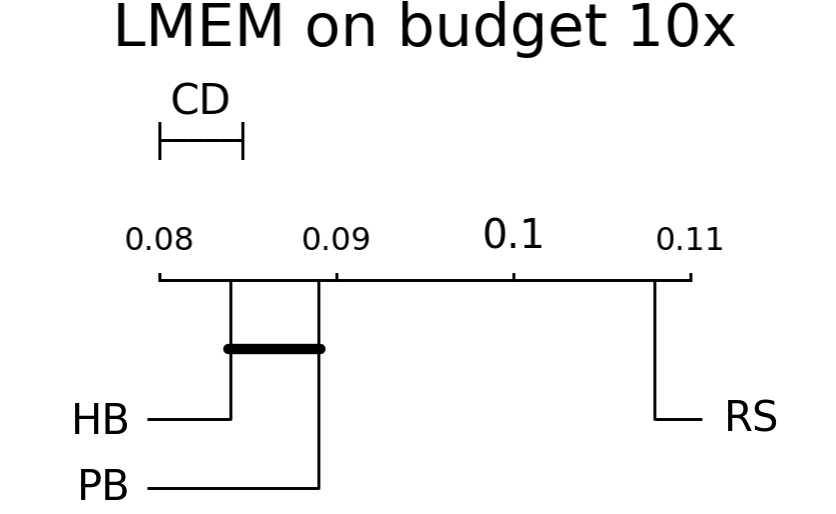} \\
    \end{tabular}
    \caption{
    \lmems{} can Autorank: (\textit{left}) output of \autorank; (\textit{right}) output of LMEMs on the same data with the simple model: $\texttt{loss}\sim\texttt{algorithm}$.
    }
    \label{fig:lmems-autorank}
\end{figure}

Figure~\ref{fig:lmems-autorank} highlights how a simple LMEM model can be used to model the entire experimental data for the $3$ algorithms of concern
, to yield the exact conclusion as current standard practice would yield using~\autorank{}.
Here, the data seen by both setups is the same.
Note, the difference in scale of the variance. 
This can be attributed to the difference in methodologies (refer to Appendix~\ref{app:lmem-glrt}).
Figure~\ref{fig:lmems-autorank} validates the use of model-based testing using LMEMs for the given data.

\subsection{Sanity checks} \label{sec:sanity}

Given that most HPO benchmarking runs will share a common set of metadata\footnote{typically: \texttt{algorithm} name, \texttt{benchmark} name, HPO \texttt{budget} spent, \texttt{seed} info, \texttt{miscellaneous} information}, LMEM models can be predefined given a data format to construct a sequence of LMEM models, that can be executed in a sequential decision-tree or workflow, checking for simple hypotheses.
These are designed to catch potential erroneous algorithms, uninformative benchmarks, or buggy results from a vast collection of benchmarking runs.
Some examples include:
\begin{inparaenum}[i)]
    \item Is any algorithm performance explained by seeds
    \item Is there any benchmark that shows no variation across algorithms 
    \item Find a complex model (hierarchically deep) that explains the data the best (see, Appendix~\ref{par:budget}).
    \note{NM: @anton which section can we refer here for the model building/selection part\\
    AG: In the appendix it would be Section \ref{par:budget} paragraph par:budget
    }
\end{inparaenum}

\begin{figure}
    \centering
       \includegraphics[width=0.55\columnwidth]{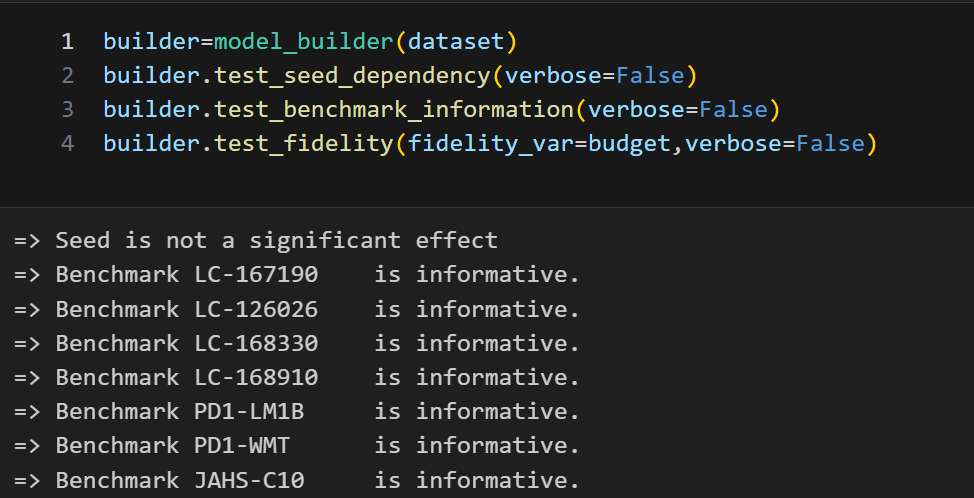} \\
    \caption{Preset sanity check are run on the experiment data to conclude that there is no algorithm where the seed explains the performance variation. There is also no benchmark where there is no performance difference across algorithms. It was also found that the used budget should be used as an interaction effect for LMEM models on this data.}
    \label{fig:lmems-sanity}
\end{figure}

Figure~\ref{fig:lmems-sanity} is an example run on the PriorBand experiment data. 
This forms an early check into the veracity of the experimentation setup and components. The Appendix \ref{app:synth-data} gives more details and verifies such preset recipes by means of synthetic datasets simulating such scenarios to catch.

\subsection{Leveraging benchmark metafeatures with significance testing}\label{sec:clustering}



\begin{figure}
    \centering
       \includegraphics[width=0.65\columnwidth]{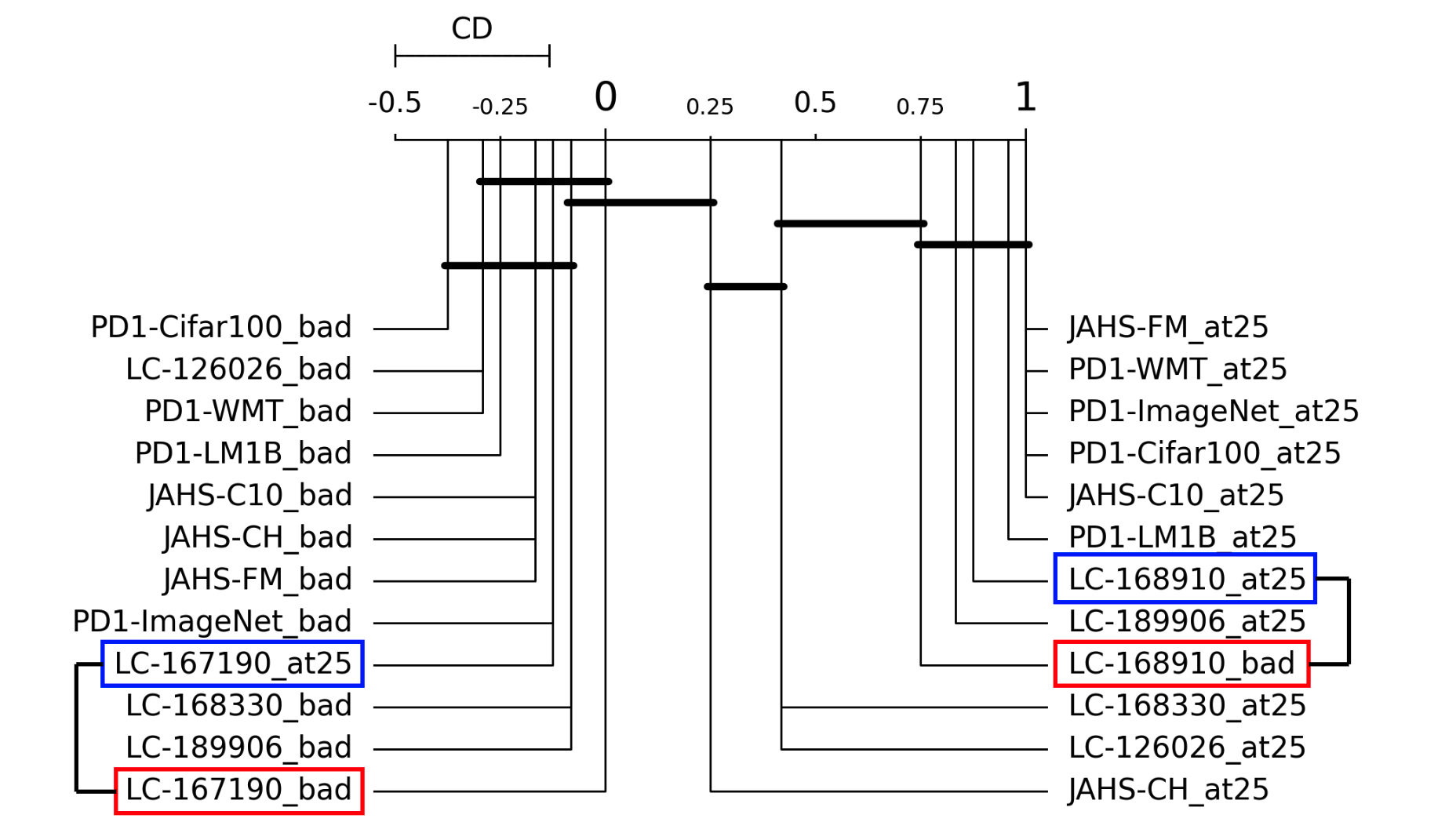} \\
    \caption{Clustering of benchmarks over prior qualities
    }
    \label{fig:case-plots}
\end{figure}

The ability of LMEMs to capture hierarchical relations in the data given an expressive model opens up the potential to analyze HPO benchmarking runs with various metafeatures in consideration.
In our case, for each benchmark, we find if \pb{} is significantly better than \hb{}, equivalent, or worse.
To achieve this, the user can select an LMEM model of their choice. 
Given this information, and the metadata information per benchmark (in this case, we know if a benchmark instance is a \textit{good} (at25) prior or a \textit{bad} prior), we test if the relative performance of \hb{} and \pb{} are significantly different or not for the two instances of the same benchmark task.
As shown in Figure~\ref{fig:case-plots}, there is significant performance difference between the two algorithms for the good-bad instances of each benchmark.
This post-hoc analysis reveals two anamolous benchmarks that on further investigation appears to not have a \textit{bad} enough expert prior designed (see, Figure~\ref{fig:prior-violins}).
~\citet{mallik-neurips23a} did not consider this and such post-hoc analysis could have potentially prompted the authors to either omit these 2 benchmarks from result aggregation or redesign the priors.
When benchmarking over large collection of benchmarks, we believe that such post-hoc analysis into which benchmarks contribute to the aggregated outcome is important for complete empirical understanding of results.

\note{
NM: @xavier, how do we write about ANOVA or~\citet{dror-tacl17a} here?\\
XB: Looking more into the difference between LMEM and ANOVA I realize that
    ANOVA can be seen as a particular case of LMEM where models are simply means.
    I'm not sure it is worth mentioning unless we could compare results with using
    ANOVA instead.
}

\subsection{Further extensions and applications}

\begin{figure}
    \centering
    \begin{tabular}{ccc|c}
        \includegraphics[width=0.22\columnwidth]{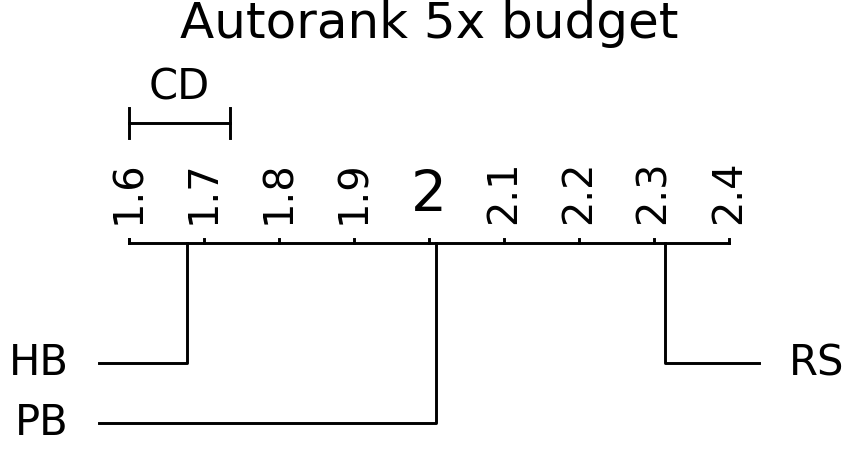} &
        \includegraphics[width=0.22\columnwidth]{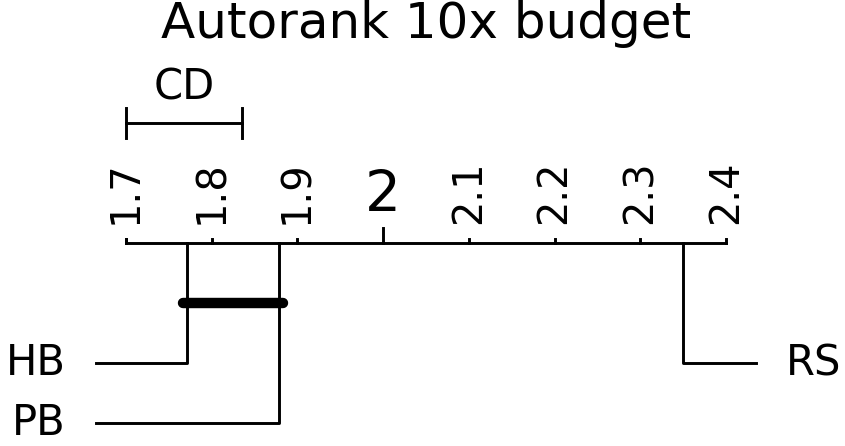} &
        \includegraphics[width=0.22\columnwidth]{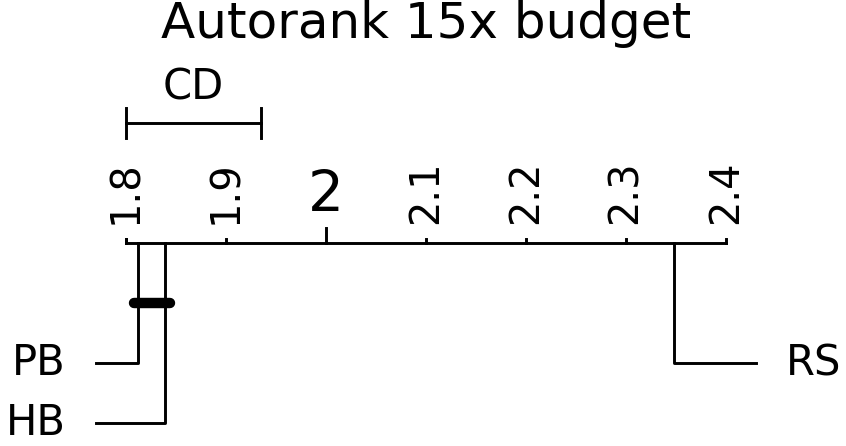}  & 
        \includegraphics[width=0.25\columnwidth]{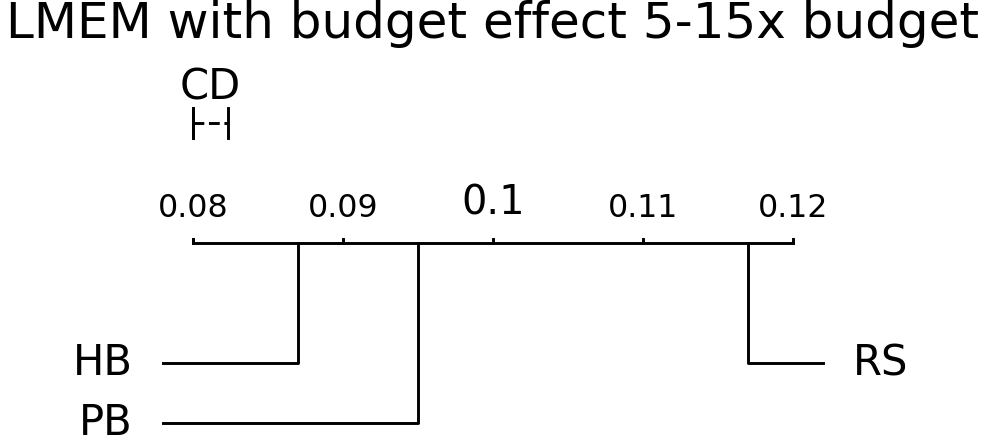} \\
    \end{tabular}
    \caption{
    (\textit{left}) \autorank{} at three different HPO budget horizons ($5\times$, $10\times$, $15\times$); (\textit{right}) LMEM trained on all available data from $5-15\times$ budget, including the budget as a random effect: $\texttt{loss}\sim\texttt{algorithm}+(1|\texttt{budget})+(1|\texttt{benchmark})$.
    }
    \label{fig:compare-anytime}
\end{figure}

The analysis from the previous section can be further expanded to a set of metafeatures.
Forward-selection could determine the metafeature(s) that when included as a random effect(s) explains the experiment data the best.
Looking into these subsets can provide useful information.
Similarly, using the HPO budget spent as an effect can bring out nuances of how algorithms perform given an HPO budget window.
We explore this briefly in Figure~\ref{fig:compare-anytime} but note that it still requires work to be the go-to analysis choice for \textit{anytime} performance.



\note{NM: do we call out the limitation here already regarding the compressed anytime performance?
and why HB>PB\\
AG: Here or in limitations?}




\section{Conclusion} \label{sec:conclusion}

In this work, we propose and demonstrate the flexible application of LMEM-based significance testing in the context of HPO Benchmarking.
More specifically, we show how LMEMs allow the modeling of potential hierarchical patterns in the benchmarking data, by accounting for random effects that can be attributed to different benchmark problems. We also show case how the HPO budget could be modeled as a fixed effect and allow for a novel compressed anytime performance analysis using a single CD plot.
Moreoever, LMEMs offer the use of metafeatures for deeper analysis into relative performance of algorithms on subsets of benchmarks.
LMEM-based methods such as these offer both the HPO researcher and practitioner to construct diverse hypotheses and obtain a different perspectives on their experiment data.
Given standard data, our open-sourced Python package offers off-the-shelf recipes for the analysis presented in this work.




\section{Limitations}

In this paper, we aim to demonstrate the rich potential of applying flexible LMEM-based testing on standard HPO benchmarking data.
We propose using these to supplement and aid existing analysis methods.
To be the gold standard of significance testing for HPO, this warrants a longer and more scientific discussion.

While linear mixed-effects models (LMEMs) offer a powerful and flexible approach, their training and testing can be significantly slower than standard methods, especially for complex models or large datasets. This is partly due to our current implementation using the \texttt{pymer4} package, which relies on R in the background. We're actively working on a full Python implementation to address both speed and user-friendliness.

However, the flexibility of LMEMs also comes with a potential downside: the ease of modifying formulas can lead to misuse by inexperienced users. To mitigate this, we recommend using only basic formulas, employing the GLRT (Generalized Likelihood Ratio Test) to validate new effects, consulting with statisticians for complex models, and fully reporting all details when using LMEMs for significance testing in publications.
It's important to acknowledge that the complexity of the model fit and potential limitations in the available data might lead to incorrectly rejecting the null hypothesis (essentially, finding a false difference). When in doubt, a simple ANOVA
may be initially preferred.

Relatedly, we lack a statistical power analysis and did not extend the application of LMEM-based
significance testing further than what is proposed by \cite{riezler-22a} to include
any control over Type II error. Further guidelines could be provided
based on works such as the one of \cite{matuschek2017balancing}.


\section{Broader Impact}

This paper applies an existing method in a post-hoc manner on existing benchmarking runs and, therefore, has no direct impact beyond the interested HPO researchers and practitioners. 
However, there is a possibility that the insights gained through the contributions of this paper can lead to more directed experiments, potentially saving computational resources and energy.





%





\begin{acknowledgements}
FH is a Hector Endowed Fellow at the ELLIS Institute T\"{u}bingen.
AG, NM, DS and FH acknowledge funding by the state of Baden-W\"{u}rttemberg through bwHPC, the German Research Foundation (DFG) through grant numbers INST 39/963-1 FUGG and 417962828, and the European Union (via ERC Consolidator Grant Deep Learning 2.0, grant no.~101045765), TAILOR, a project funded by EU Horizon 2020 research and innovation programme under GA No 952215. Views and opinions expressed are however those of the author(s) only and do not necessarily reflect those of the European Union or the European Research Council. Neither the European Union nor the granting authority can be held responsible for them.

\begin{center}\includegraphics[width=0.3\textwidth]{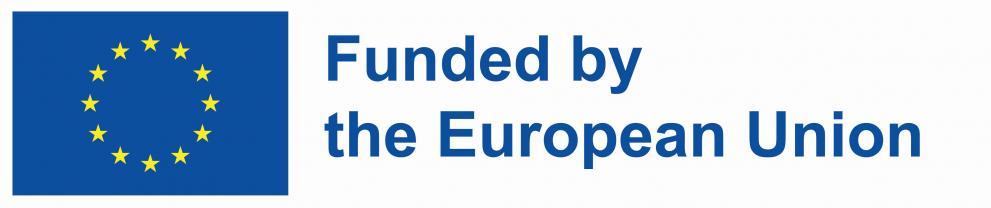}\end{center}

\end{acknowledgements}

\newpage


\bibliography{bib/lib,bib/local,bib/proc,bib/strings}

\newpage


\section*{Submission Checklist}

\begin{enumerate}
\item For all authors\dots
  \begin{enumerate}
  \item Do the main claims made in the abstract and introduction accurately
    reflect the paper's contributions and scope?
    %
    \answerYes{}
  \item Did you describe the limitations of your work?
    %
    \answerYes{}
  \item Did you discuss any potential negative societal impacts of your work?
    %
    \answerNA{}
  \item Did you read the ethics review guidelines and ensure that your paper
    conforms to them? \url{https://2022.automl.cc/ethics-accessibility/}
    %
    \answerYes{}
  \end{enumerate}
\item If you ran experiments\dots
  \begin{enumerate}
  \item Did you use the same evaluation protocol for all methods being compared (e.g.,
    same benchmarks, data (sub)sets, available resources)?
    %
    \answerYes{}
  \item Did you specify all the necessary details of your evaluation (e.g., data splits,
    pre-processing, search spaces, hyperparameter tuning)?
    %
    \answerNA{}
  \item Did you repeat your experiments (e.g., across multiple random seeds or splits) to account for the impact of randomness in your methods or data?
    %
    \answerNA{}
  \item Did you report the uncertainty of your results (e.g., the variance across random seeds or splits)?
    %
    \answerNA{}
  \item Did you report the statistical significance of your results?
    %
    \answerNA{}
  \item Did you use tabular or surrogate benchmarks for in-depth evaluations?
    %
    \answerNA{}
  \item Did you compare performance over time and describe how you selected the maximum duration?
    %
    \answerNA{}
  \item Did you include the total amount of compute and the type of resources
    used (e.g., type of \textsc{gpu}s, internal cluster, or cloud provider)?
    \answerNA{}
  \item Did you run ablation studies to assess the impact of different
    components of your approach?
    \answerNA{}
  \end{enumerate}
\item With respect to the code used to obtain your results\dots
  \begin{enumerate}
\item Did you include the code, data, and instructions needed to reproduce the
    main experimental results, including all requirements (e.g.,
    \texttt{requirements.txt} with explicit versions), random seeds, an instructive
    \texttt{README} with installation, and execution commands (either in the
    supplemental material or as a \textsc{url})?
    \answerNo{The URL will be released as part of the non-anonymized version of the paper.}
  \item Did you include a minimal example to replicate results on a small subset
    of the experiments or on toy data?
    \answerYes{}
  \item Did you ensure sufficient code quality and documentation so that someone else
    can execute and understand your code?
    \answerYes{}
  \item Did you include the raw results of running your experiments with the given
    code, data, and instructions?
    \answerNA{}
  \item Did you include the code, additional data, and instructions needed to generate
    the figures and tables in your paper based on the raw results?
    \answerYes{}
  \end{enumerate}
\item If you used existing assets (e.g., code, data, models)\dots
  \begin{enumerate}
  \item Did you cite the creators of used assets?
    \answerYes{}
  \item Did you discuss whether and how consent was obtained from people whose
    data you're using/curating if the license requires it?
    \answerYes{}
  \item Did you discuss whether the data you are using/curating contains
    personally identifiable information or offensive content?
    \answerNA{}
  \end{enumerate}
\item If you created/released new assets (e.g., code, data, models)\dots
  \begin{enumerate}
    \item Did you mention the license of the new assets (e.g., as part of your code submission)?
    \answerYes{}
    \item Did you include the new assets either in the supplemental material or as
    a \textsc{url} (to, e.g., GitHub or Hugging Face)?
    \answerNA{}
  \end{enumerate}
\item If you used crowdsourcing or conducted research with human subjects\dots
  \begin{enumerate}
  \item Did you include the full text of instructions given to participants and
    screenshots, if applicable?
    \answerNA{}
  \item Did you describe any potential participant risks, with links to
    Institutional Review Board (\textsc{irb}) approvals, if applicable?
    \answerNA{}
  \item Did you include the estimated hourly wage paid to participants and the
    total amount spent on participant compensation?
    \answerNA{}
  \end{enumerate}
\item If you included theoretical results\dots
  \begin{enumerate}
  \item Did you state the full set of assumptions of all theoretical results?
    \answerNA{}
  \item Did you include complete proofs of all theoretical results?
    \answerNA{}
  \end{enumerate}
\end{enumerate}

\newpage
\appendix

\newpage


\appendix
\section{LMEMs and GLRTs overview} \label{app:lmem-glrt}

\note{AM: Decide on the short or long version.}
\note{NM: Could we cite the main tutorial and link the blog here?}
\note{AM: That would be \citet{hagmann-arxiv23a}. What blog do you mean?}

\subsection{Short version}\label{sec:short_math}
This likelihood-based testing approach trains an LMEM on the experimental data and estimates the mean of each algorithm based on the model's parameters. The results are normal distributions that can be compared via statistical tests, here we employ pairwise Tukey HSD tests. 
The model itself is used to capture additional information, disregarded in classical testing. We include the respective benchmark, seed, fidelity, and meta-features like the prior quality of each result. LMEMs are especially fit for this use case as they model inherent hierarchies in the data through the use of \textit{random effects}. A variable like benchmark as \textit{random effect} is assumed to be not fully observed but a sample from a zero-mean random distribution with an internally estimated variance-covariance matrix. As a result, they account for any variation caused by e.g. a different benchmark, effectively unifying the data. The approach therefore seeks to produce general results beyond what statistical tests normally could.\\
Additionally, two LMEMs $M_0$ and $M_1$ (with likelihoods $l_0$ and $l_1$ and numbers of parameters $k_0$ and $k_1$) can function as representations of hypotheses to be compared with the Generalized Likelihood Ratio Test (GLRT). As the models are fit by maximum likelihood, their likelihoods are normally distributed, producing a $\chi^2$-testing distribution.
\begin{equation*}\label{eq:glrt}
    2 \log\frac{l_0}{l_1} \sim \chi^2_{k_0-k_1}
\end{equation*}



\subsection{Linear Mixed Effect Models}

This work showcases the use of Linear Mixed Effect Models (LMEMs) for a model-based significance testing approach. These models have long been in use for their effectiveness at handling grouped data and within-group correlation, which is why they are now applied to HPO settings \citep{pinheiro-springer00a}. Briefly explained LMEMs extend the simple Linear Model (\eqref{linear model}) by additional components.

\begin{equation}
    \vec{Y} = \mat{X} \boldsymbol{\beta} + \boldsymbol{\epsilon} \text{,\quad where } \boldsymbol{\epsilon}\sim \mathcal{N}(0,\boldsymbol{\Lambda}_{\theta}) \label{linear model}
\end{equation}

In LMEMs, the vector $\vec{\beta}$ is called \textit{fixed effects} to contrast the second, \textit{random effects}-vector $\vec{b}$, which has its own design-matrix $\mat{Z}$. 

\begin{equation}
    \vec{Y} = \mat{X} \boldsymbol{\beta} +\mat{Z}\vec{b} + \boldsymbol{\epsilon}
\end{equation}
Contrary to $\boldsymbol{\beta}$, the random vector is assumed to consist only of samples from a normal distribution, similar to the error vector, instead of fully and reliably observed values, such that:
\begin{equation}
    \vec{b}\sim \mathcal{N}(0,\boldsymbol{\psi}_{\theta})
\end{equation}
LMEMs now optimize the fixed effects matrix $\mat{X}$ directly, while the random effects matrix $\mat{Z}$ is determined from the distribution of $\vec{b}$, by estimating its variance-covariance matrix $\boldsymbol{\psi}_{\theta}$ through likelihood-maximization.\newline
Now assume a model using the algorithm as fixed and the benchmark as a random effect (\eqref{eq:lmem_formula}). It has a grand intercept $\mu$ and for each algorithm $a$ an effect $\nu_{a}$ that gets activated by its indicator function $\mathbb{I}_{a}$. The same counts for the random effect benchmark, with its effects $\nu_{b}$ and indicator function $\mathbb{I}_{b}$.
\begin{equation} \label{eq:lmem_formula}
    Y= \mu + \sum_{a \in algorithms} \nu_{a}*\mathbb{I}_{a} + \sum_{b \in benchmarks} \nu_{b}*\mathbb{I}_{b} +\epsilon_{residual}
\end{equation}
This is logically extended to an ordinal variable $x$, where the sum over indicator functions is replaced by $\nu_{x}*x$ and interaction effects, with double sums and $\nu_{xy}*\mathbb{I}_{x \land y}$. This can introduce levels of hierarchy, which is an essential use case of LMEMs. As an example, for a categorical meta-parameter (or fidelity) $p$ we can introduce a fixed effect $\nu_{p}$ and an interaction effect $\nu_{ap}$ to retrieve individual performances under each value of $p$:
\begin{equation}
    Y= \mu + \nu_{p} \sum_{a \in algorithms} \left(\nu_{a}*\mathbb{I}_{a} + \sum_{p \in parameter} \nu_{ap}*\mathbb{I}_{p} \right) +\epsilon_{residual}
\end{equation}

\subsection{Estimated Marginal Means and Tukey-HSD}\label{sec:ems}

\paragraph{Estimated Marginal Means} Having built the model, we use \textit{Estimated Marginal Means} (EMMs) to generate the testing data. In this process, an EMM grid is generated containing all unique values of all effects. Over this EMM grid, we can calculate the mean and standard error for each algorithm, again using the variance-covariance matrix $\boldsymbol{\psi}_{\theta}$. The resulting means and standard errors are then used as the basis for the actual significance test.\newline
This is a unique property of model-based testing methods because we do not test on the data itself but rather use it as response variable for the model. This has considerable benefits, as it alleviates any restrictions on the distributional properties of the data like being normally distributed or having homogeneous variances. Any data can train the model while the model parameters themselves asymptotically follow a normal distribution, as they have been obtained by maximizing the likelihood \citep{demsar-06a}.

\paragraph{Pairwise comparisons with Tukey HSD} This understanding allows for a simple t-test for pairwise comparisons of each algorithm. As recommended by \citet{riezler-22a} we use the \textit{Tukey HSD test} to control the per-experiment Type-I error, as it is well suited for larger numbers of algorithms, contrary to the \textit{Bonferroni correction}. The Tukey HSD test is a multiple testing corrected t-test introduced by \citet{tukey-49a}. We obtain the test statistic from the distance of two algorithms' estimated means and their common standard error:
\begin{equation}
    q_s=\frac{|m_1-m_2|}{SE_{m_1,m_2}}
\end{equation}
The critical value $q^*$ of this distribution is retrieved from the \textit{studentized range distribution}, depending on the elected Type I error rate, the number of algorithms $k$ and the number of observations per algorithm $N_a$, which give the degrees of freedom per algorithm as $df_a=N_a-k$. Using the cumulative studentized range distribution we can determine the p-value for this comparison of algorithms.

\paragraph{CD-Diagrams} From the Tukey HSD test we obtain an individual p-value for each comparison. This is ideal for constructing a CD-Diagram, a method introduced by \citet{demsar-06a}, using the Estimated Means as algorithm ranks. Importantly, we also obtain an individual \textit{Critical Difference} for each comparison, which is why we extended the CD-Diagram to show the range from smallest to largest Critical Difference above.

\subsection{Generalized Likelihood Ratio Testing}

These LMEMs can in principle be extended by any possible factor but this comes at the cost of complexity and ability to generalize. To optimize this trade-off, we follow the approach of Riezler et al. to employ the \textit{Generalized Likelihood Ratio Test} (GLRT) to compare models, specifically one with ($M_1$) against one without ($M_0$) the effect in question \citep{riezler-22a}. The GLRT compares the two models' likelihood $l_1$ and $l_0$ (Equation \ref{eq:glrt}) which in turn are both approximately normal, as they come from a maximum likelihood optimization. Therefore their ratio is $\chi ^2$ distributed with $k_0-k_1$ degrees of freedom for $k_i$ parameters of model $M_i$ \citep{riezler-22a}.
\begin{equation*}\label{eq:glrt}
    2 \log\frac{l_0}{l_1} \sim \chi^2_{k_0-k_1}
\end{equation*}
Using the critical value from this distribution, we can decide on which effects to add per the significance of the resulting p-value of the GLRT. A significant p-value lets us reject the null hypothesis that both models perform the same and, when additionally $l_0>l_1$, conclude that the effect increases the model's power significantly.
\section{Data from \pb{}} \label{app:pb-data}

In this section, we refer to all the relevant information from the original \pb{} work~\citep{mallik-neurips23a} that supplements our case study of applying LMEM-based significant testing.

\subsection{Subset considered}

\begin{figure}[htpb]
\begin{centering}
    \includegraphics[scale=0.3]{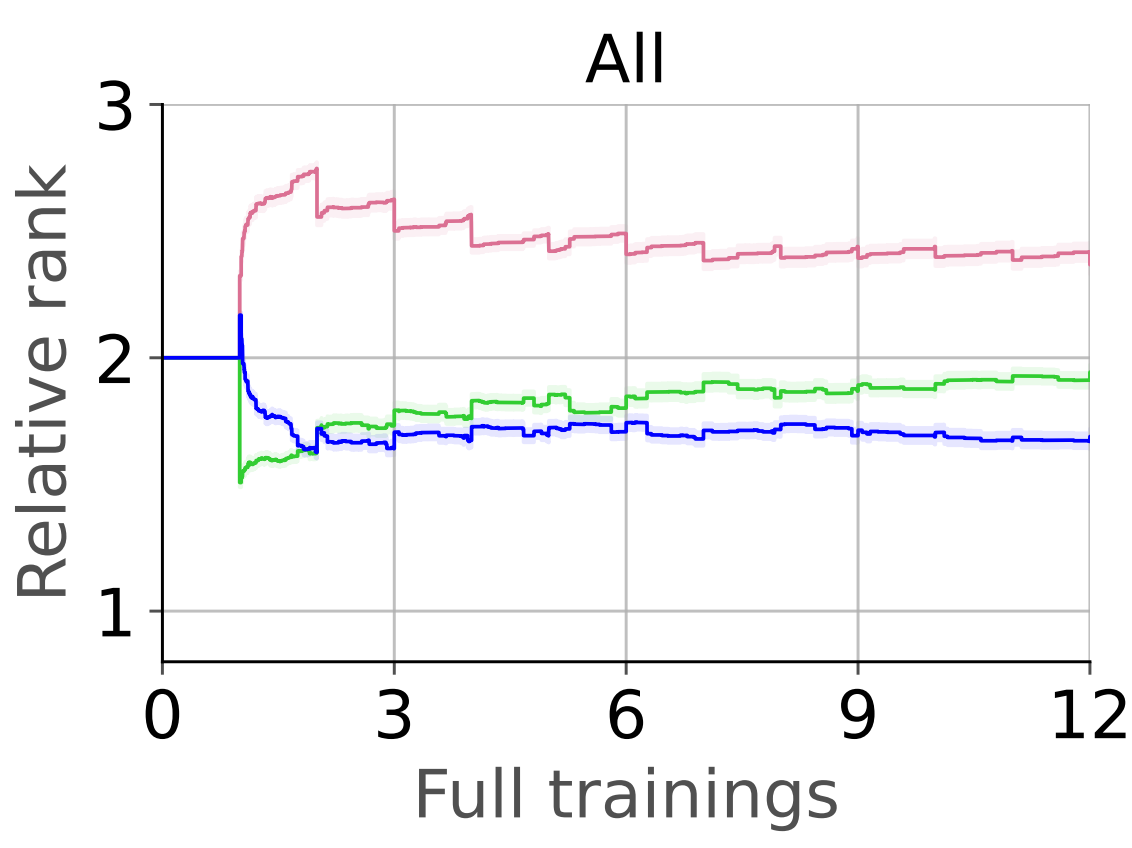}
    \includegraphics[scale=0.3]{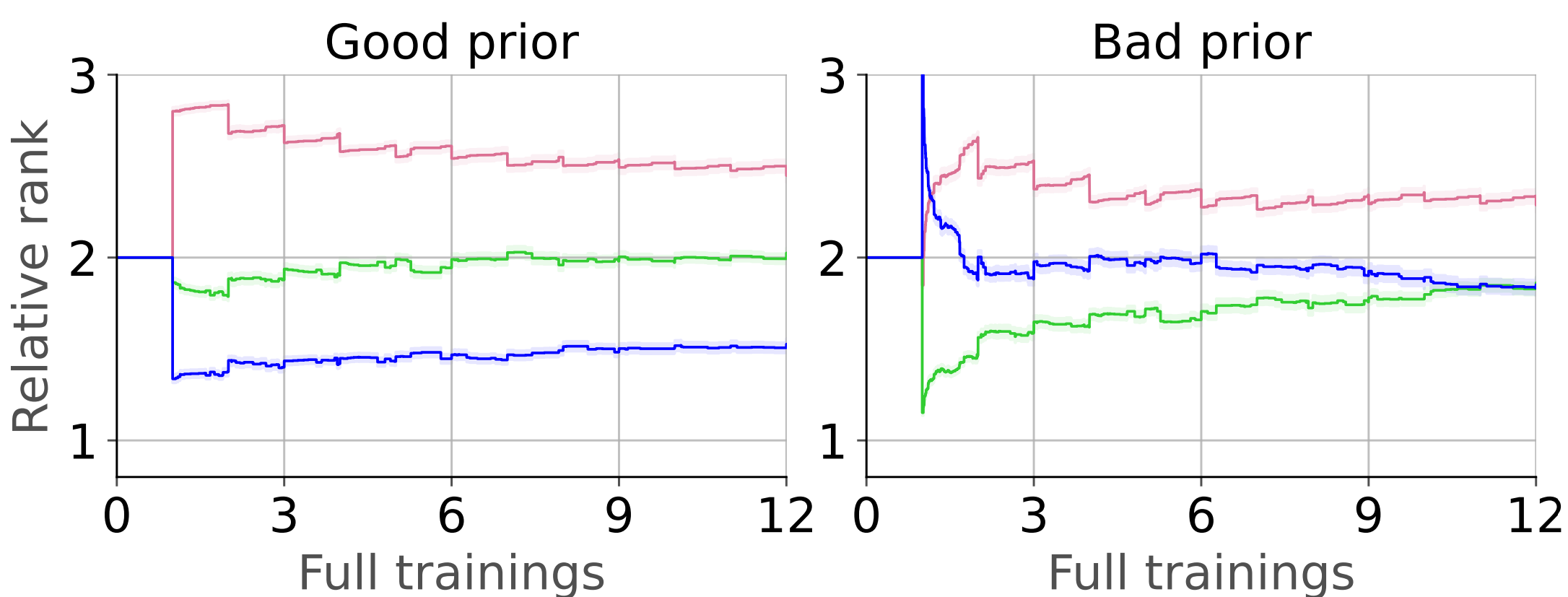}
    \includegraphics[scale=0.28]{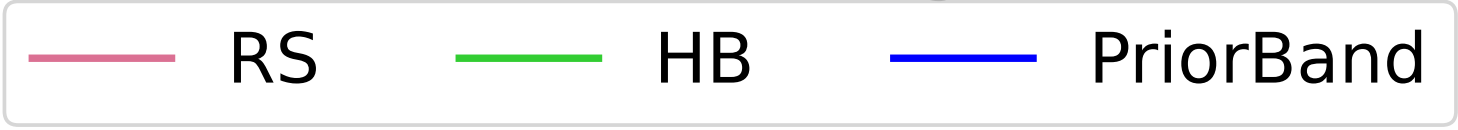}
    \caption[PriorBand's Relative Ranks]{\textbf{Relative Ranks plot from the PriorBand Paper:} On the top: Relative ranks under both good (at25) and bad priors combined. On the bottom: Performance under individual priors. (Figure sourced from ~\citet{mallik-neurips23a})}
    \label{fig:PB_rel_rank}
\end{centering}
\end{figure}

Figure~\ref{fig:PB_rel_rank} shows the primary result of concern in our case study.
We want to understand the role of the different benchmarks, the quality of the default prior for that benchmark and how \hb{} and \pb{} differ in performance on them.

\subsection{Metadata of experiment data}

\begin{figure}[htpb]
\begin{centering}
    \includegraphics[width=0.8\textwidth]{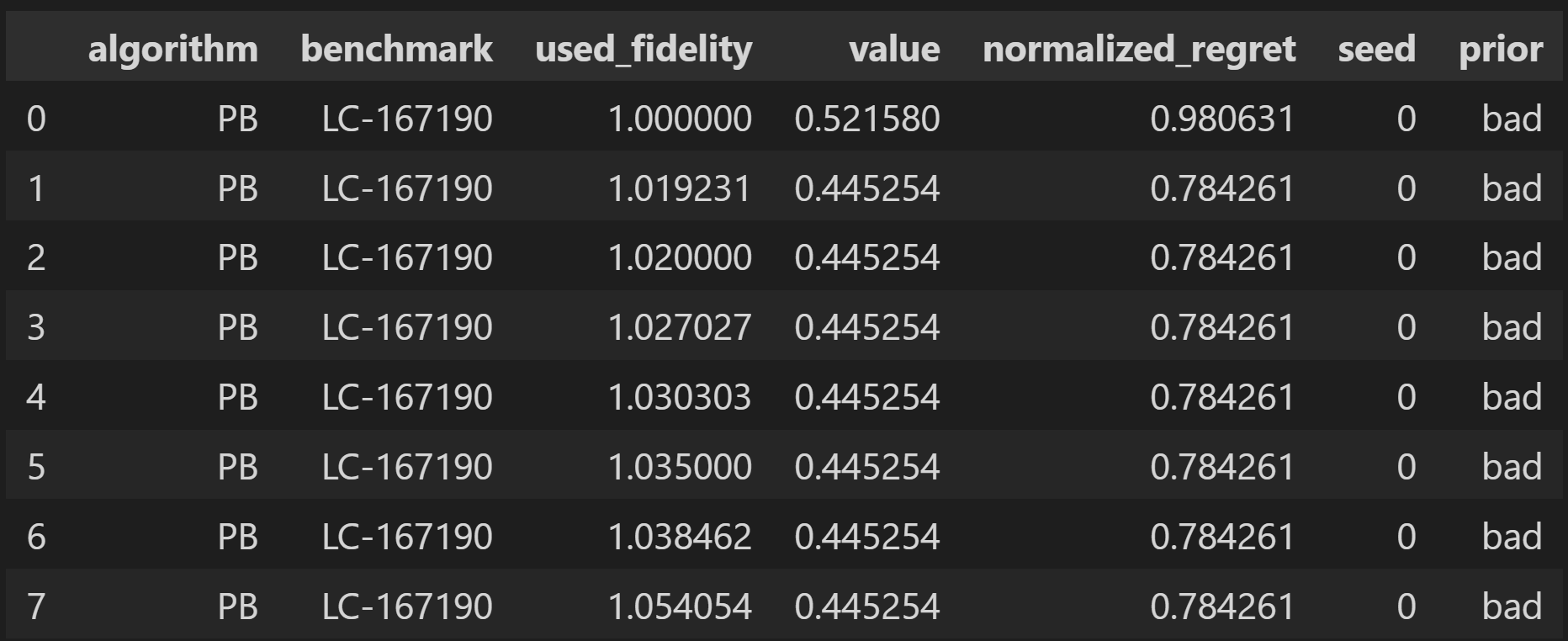}
    \caption[PriorBand's data]{\textbf{Excerpt from the PriorBand paper's data}  \citep{mallik-neurips23a}\\
    The data contains information on the loss and algorithm and on which benchmark, seed, fidelity, and prior it was acquired on.}
    \label{fig:head_full}
\end{centering}
\end{figure}

\begin{figure}[htpb]
\begin{centering}
    \includegraphics[width=0.45\textwidth]{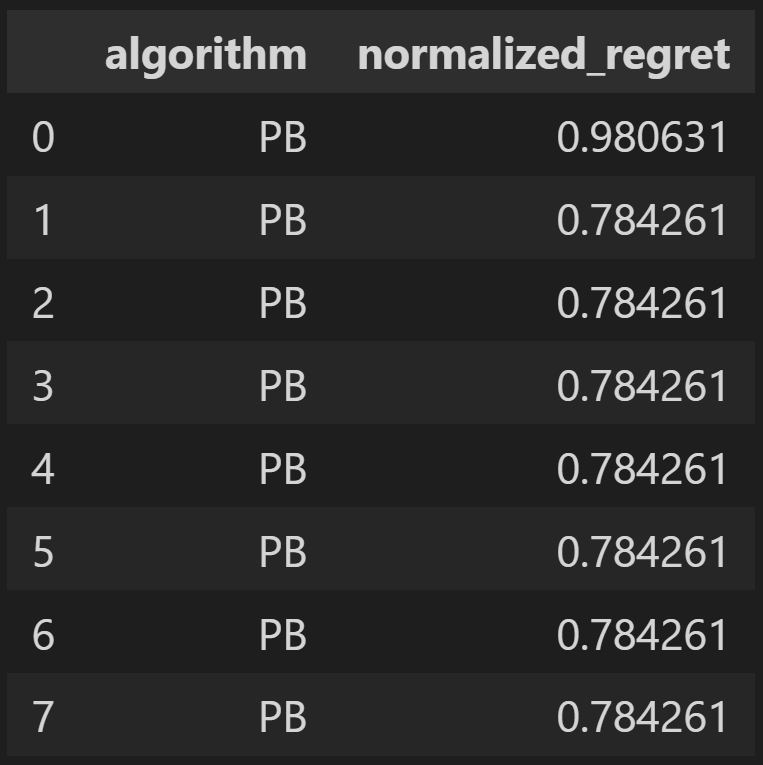}
    \caption[PriorBand's data]{\textbf{Data used for comparing \lmems{} to \autorank{}}\\
    The data contains the loss and the algorithm.}
    \label{fig:head_autorank}
\end{centering}
\end{figure}

\begin{figure}[htpb]
\begin{centering}
    \includegraphics[width=0.7\textwidth]{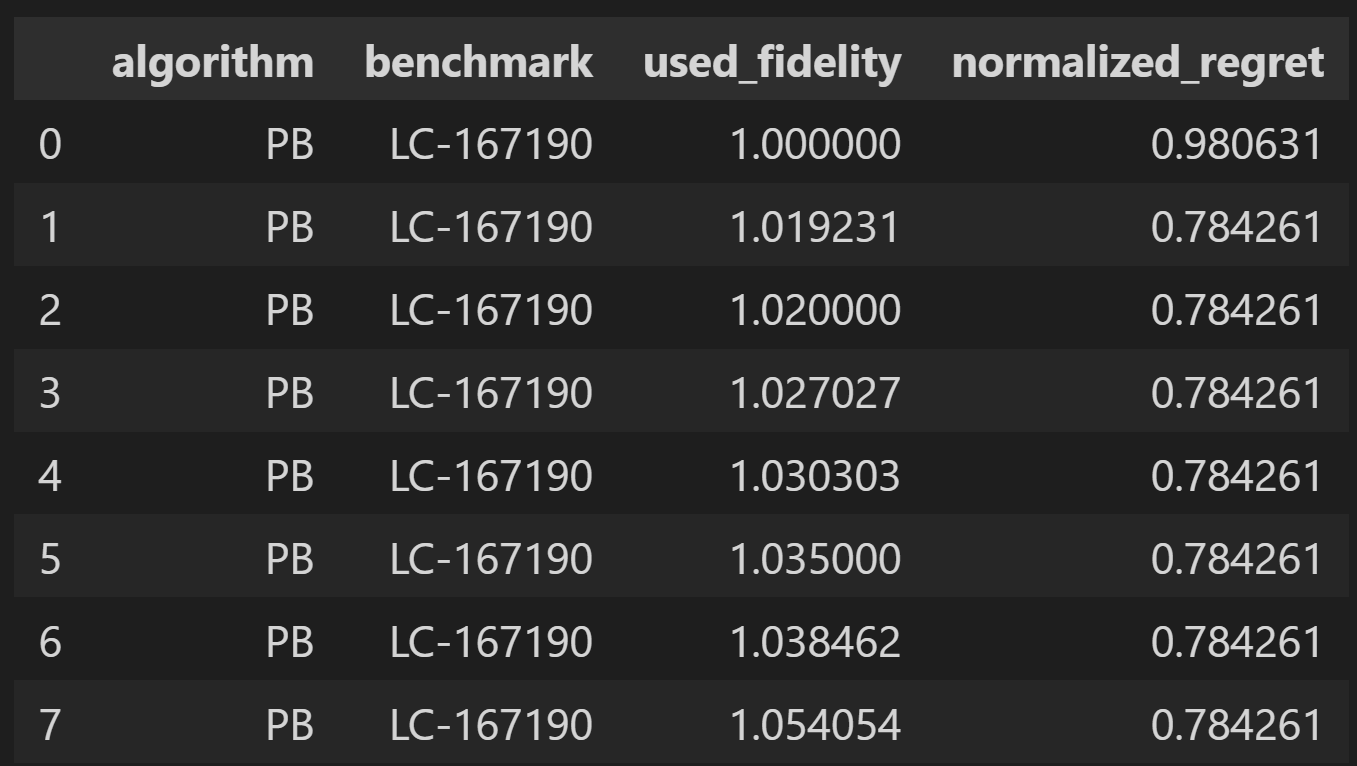}
    \caption[PriorBand's data]{\textbf{Data used for anytime analysis}\\
    The data contains the loss, algorithm and training budget.}
    \label{fig:head_budget}
\end{centering}
\end{figure}

\begin{figure}[htpb]
\begin{centering}
    \includegraphics[width=0.75\textwidth]{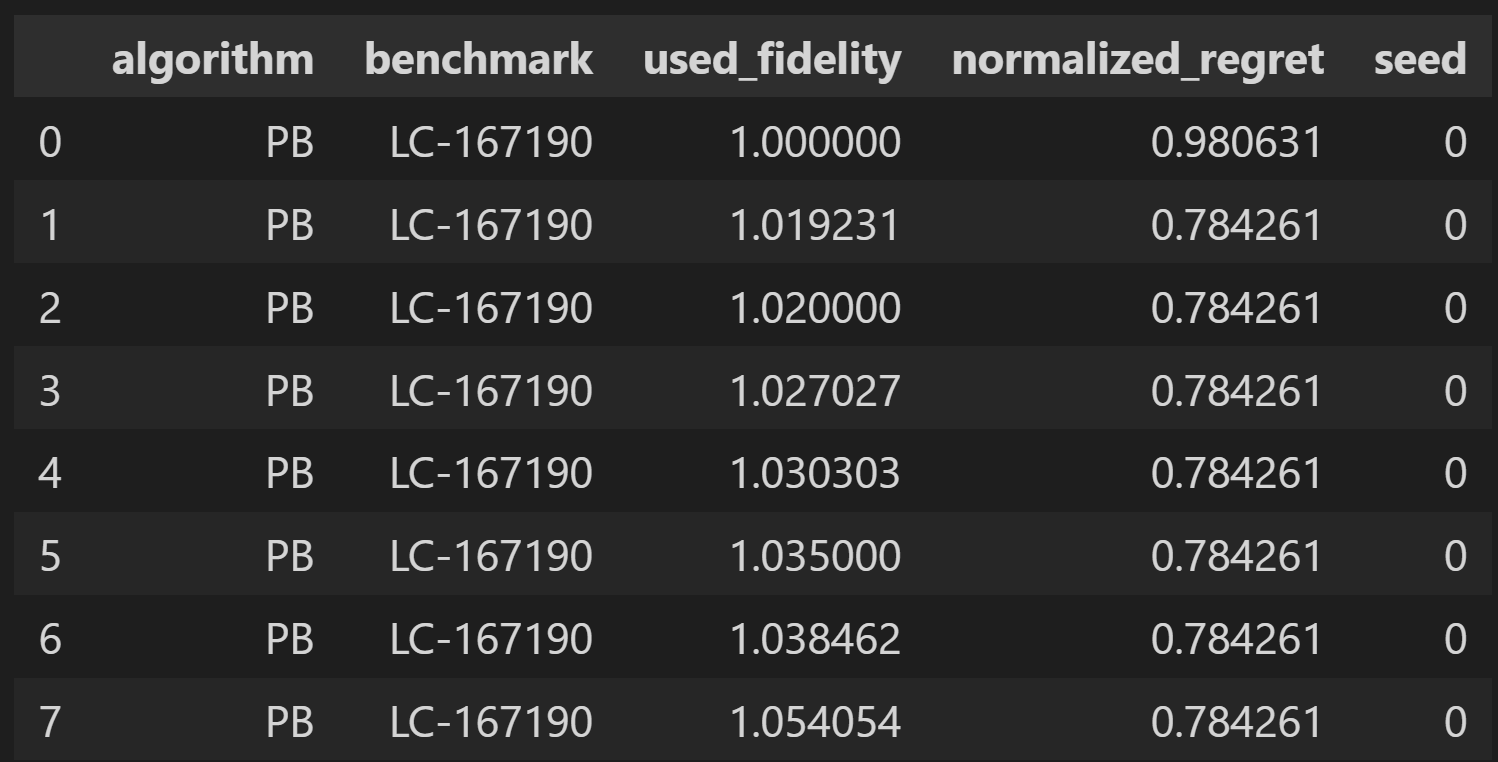}
    \caption[PriorBand's data]{\textbf{Data used in the sanity checks}\\
    The data contains information on the loss and algorithm and on which benchmark, seed, and fidelity it was acquired on.}
    \label{fig:head_sanity}
\end{centering}
\end{figure}

\begin{figure}[htpb]
\begin{centering}
    \includegraphics[width=0.8\textwidth]{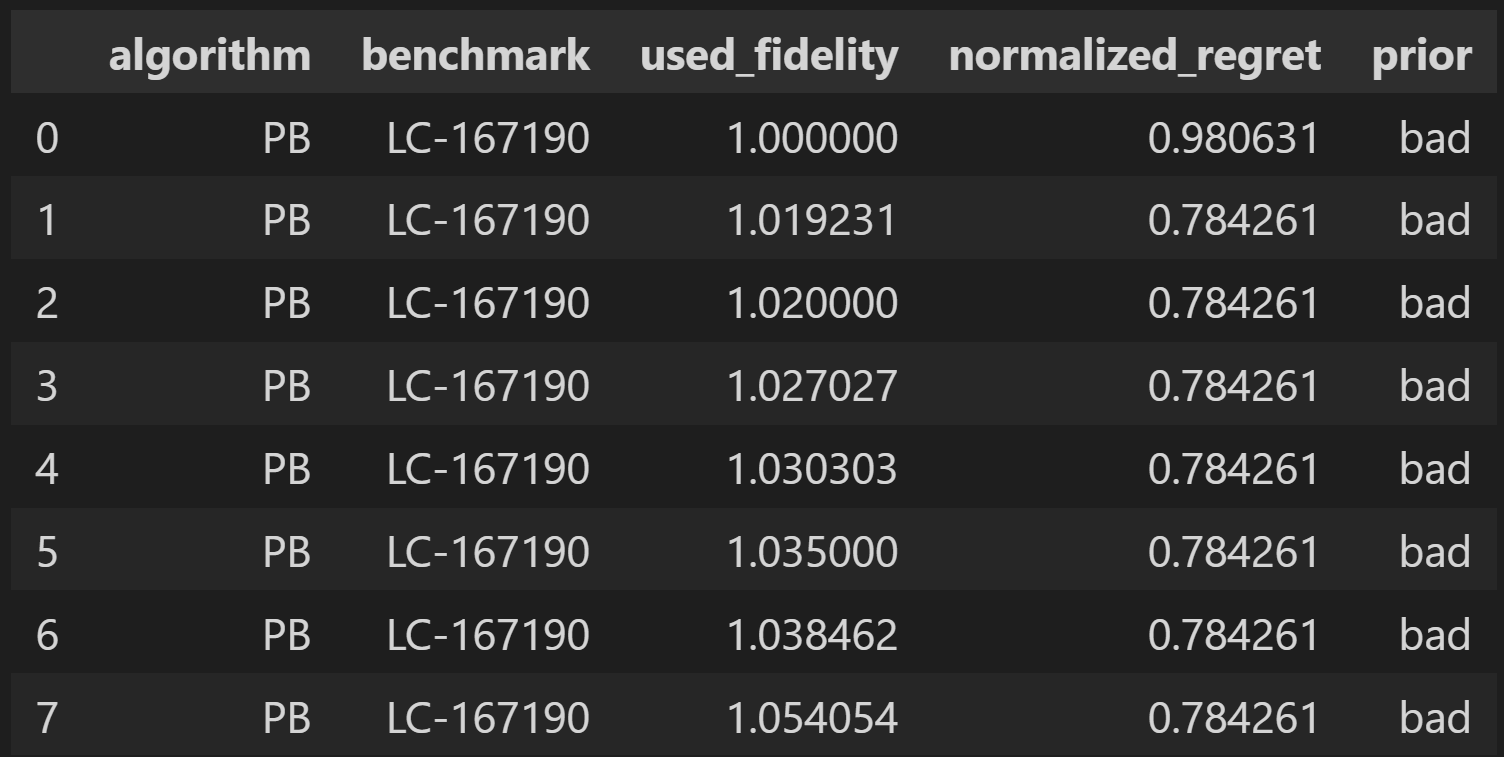}
    \caption[PriorBand's data]{\textbf{Data used for benchmark clustering}\\
    The data contains information on the loss and on which benchmark, fidelity, and prior it was acquired on.}
    \label{fig:head_clustering}
\end{centering}
\end{figure}

In this section we illustrate the structure of the HPO Benchmarking data we inherited. The original data from \citep{mallik-neurips23a} is shown in Figure \ref{fig:head_full}. For Section \ref{sec:drop-autorank} we use only the information about the loss and algorithm (Figure \ref{fig:head_autorank}). For anytime performance analysis, we additionally include information about the training budget (Figure \ref{fig:head_budget}). In the sanity checks (Section \ref{sec:sanity}), we also include the seed to analyze for seed dependencies (Figure \ref{fig:head_sanity}). Finally in the benchmark clustering (Section \ref{sec:clustering}), we include the loss, algorithm, benchmark, budget and prior quality (Figure \ref{fig:head_clustering}).

\subsection{Supporting plots borrowed}

\begin{figure}[htbp]
    \centering
    \begin{tabular}{c}
       \includegraphics[width=0.75\textwidth]{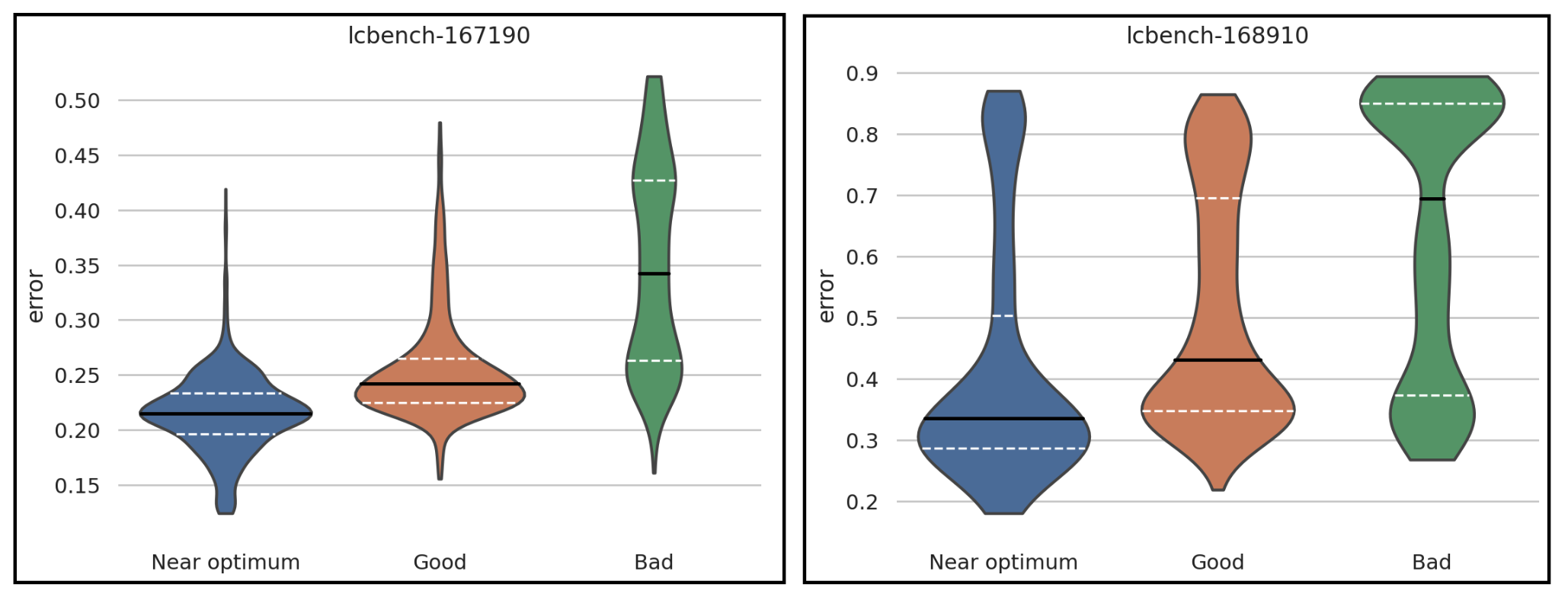} \\
    \end{tabular}
    \caption{\textbf{Violin plots from the PriorBand Paper:} Both benchmarks that showed insignificant differences in prior qualities turn out to have strongly bimodal \textit{bad} prior distributions with good quality secondary modes.}
    \label{fig:prior-violins}
\end{figure}

Figure~\ref{fig:prior-violins} shows the distribution of error given a prior configuration to sample around.
The different violins indicate different prior inputs.
These particular $2$ benchmarks from ~\citet{mallik-neurips23a} were marked in our analysis to be the odd ones under expected behaviour and trends with other benchmarks.
On looking deeper, it turned out these $2$ benchmarks have bimodal \textit{bad} prior distributions.
With one of the modes at a similar performance level as the good prior mode.
Thus, suggesting that these benchmarks could have been dropped from the aggregation results or the prior qualities should have been reassessed.

\begin{figure}
    \centering
    \includegraphics[width=0.75\textwidth]{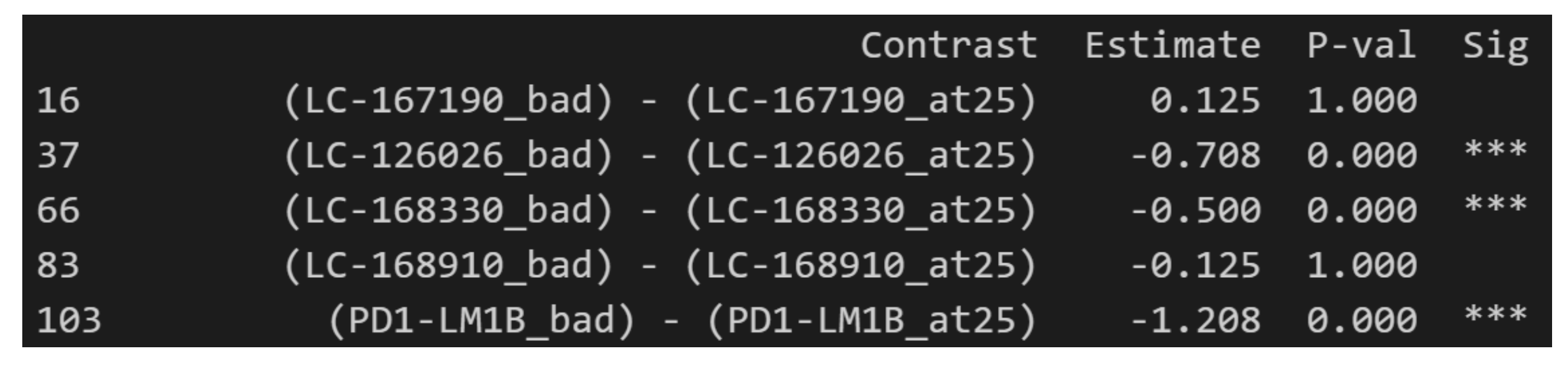}
    \caption{\textbf{Benchmark compared under good and bad priors via clustering:} Two benchmarks, \textit{LC-Bench 167190} and \textit{LC-Bench 168910} show no significant difference between their prior variants.}
    \label{fig:enter-label}
\end{figure}
       
\newpage
\section{Sanity checks on synthetic data} \label{app:synth-data}

\begin{figure}[htpb]
\begin{centering}
    \includegraphics[width=0.9\textwidth]{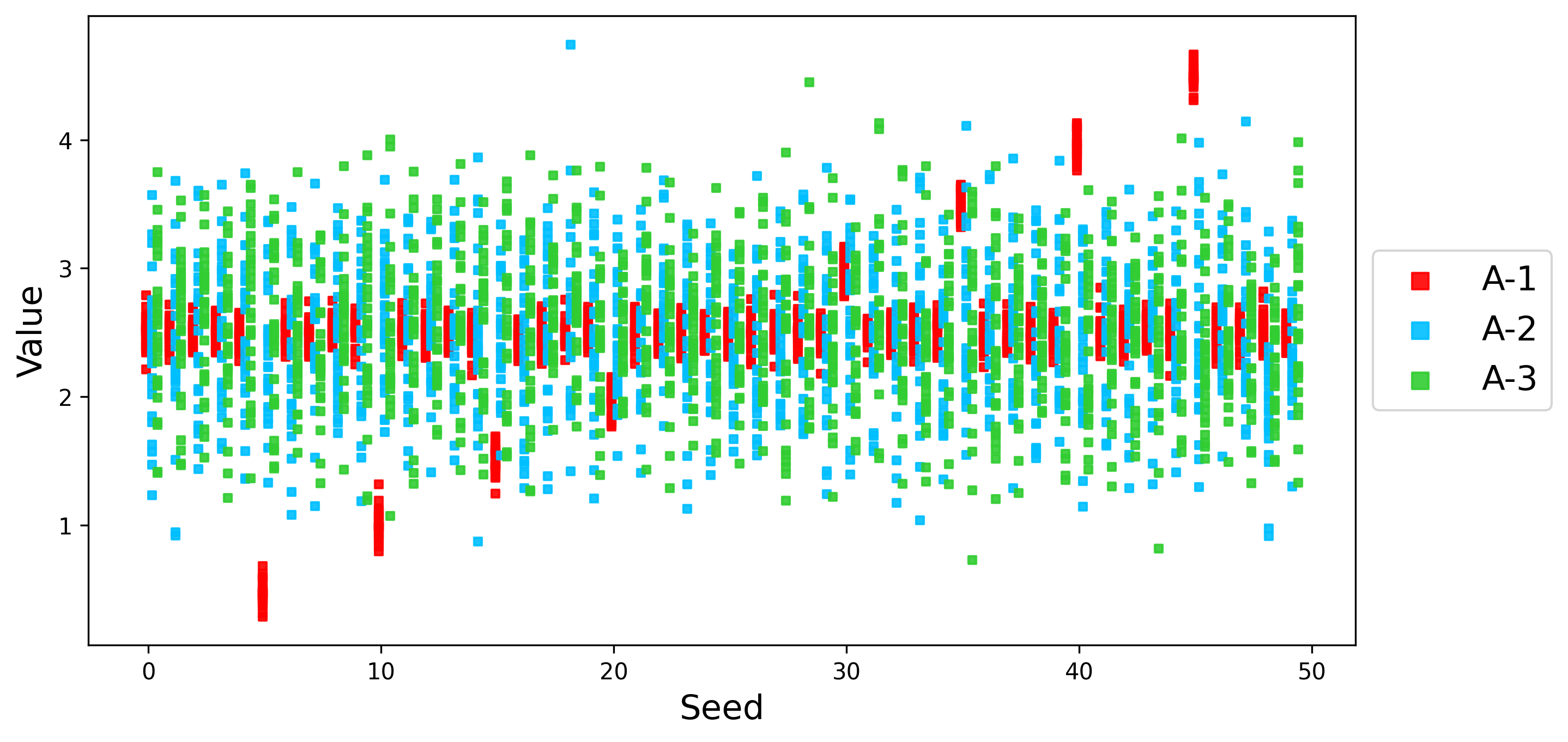}
    \caption[Dataset with Seed-dependent Algorithms]{\textbf{Seed-dependent synthetic dataset} We create a synthetic dataset with three algorithms where for some seeds algorithm A-1 uses $0.1*seed$ as mean.}
    \label{fig:seed_df}
\end{centering}
\end{figure}

In Section \ref{sec:sanity} we present several sanity checks that the PriorBand data passed. We now create synthetic datasets, to show other results of these checks.
\paragraph{Seed-Independency}\label{par:seed-dep} We have generated a small synthetic dataset of three algorithms at 50 seeds that share the same mean (2.5) and variance (0.55), but where algorithm A-1 is influenced by the seed (Figure \ref{fig:seed_df}).
A GLRT compares two models, with (Equation \ref{eq:seed_simple_model}) and without seed effect (Equation \ref{eq:seed_complex_model}), to see whether an algorithm performance might depend on the seed chosen.
\begin{equation}\label{eq:seed_simple_model}
    \textbf{loss} \boldsymbol{\sim} \textbf{algorithm}
\end{equation}
\begin{equation}\label{eq:seed_complex_model}
    \textbf{loss} \boldsymbol{\sim} \textbf{algorithm} \boldsymbol{+} \textbf{(0 + algorithm|seed)}
\end{equation}
The second model fits a random effect for each seed and algorithm. If this does not increase the significance of the model, the GLRT will reject it meaning we cannot make a significant connection between seed and algorithm performance. In this case, it accepts the complex model, so we look for large variances in this random effect. These variances show exactly which algorithm is affected.
\begin{figure}[htpb]
\begin{centering}
    \includegraphics[width=\textwidth]{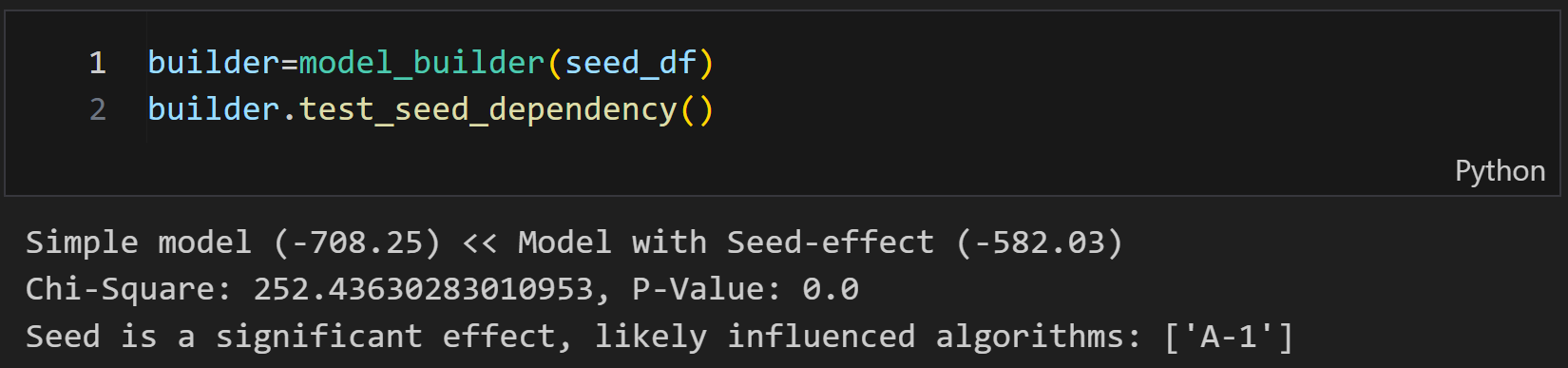}
    \caption[Code: Seed Dependency Test]{\textbf{Test on Seed dependency} We use a GLRT to compare a model that includes the seed as a factor against one that does not (Simple model). The p-value is significant and the seed model is more likely than the simple one, so the seed is a significant effect. In the second step, we analyze the effect's variances and correctly determine algorithm A-1 to be the only affected algorithm.}
    \label{fig:seed_code}
\end{centering}
\end{figure}

\paragraph{Benchmark relevance} For each benchmark, we train a model with (Equation \ref{eq:benchmark_1_model}) and without (Equation \ref{eq:benchmark_algo_model}) algorithm effect, so assuming either significant or insignificant differences between the algorithms.
\begin{equation}\label{eq:benchmark_1_model}
    \textbf{loss} \boldsymbol{\sim} \boldsymbol{1}
\end{equation}
\begin{equation}\label{eq:benchmark_algo_model}
    \textbf{loss} \boldsymbol{\sim} \textbf{algorithm}
\end{equation}
Again we show this using a test set where the algorithm's mean performance depends on the Benchmark:

\begin{figure}[htpb]
\begin{centering}
    \includegraphics[width=\textwidth]{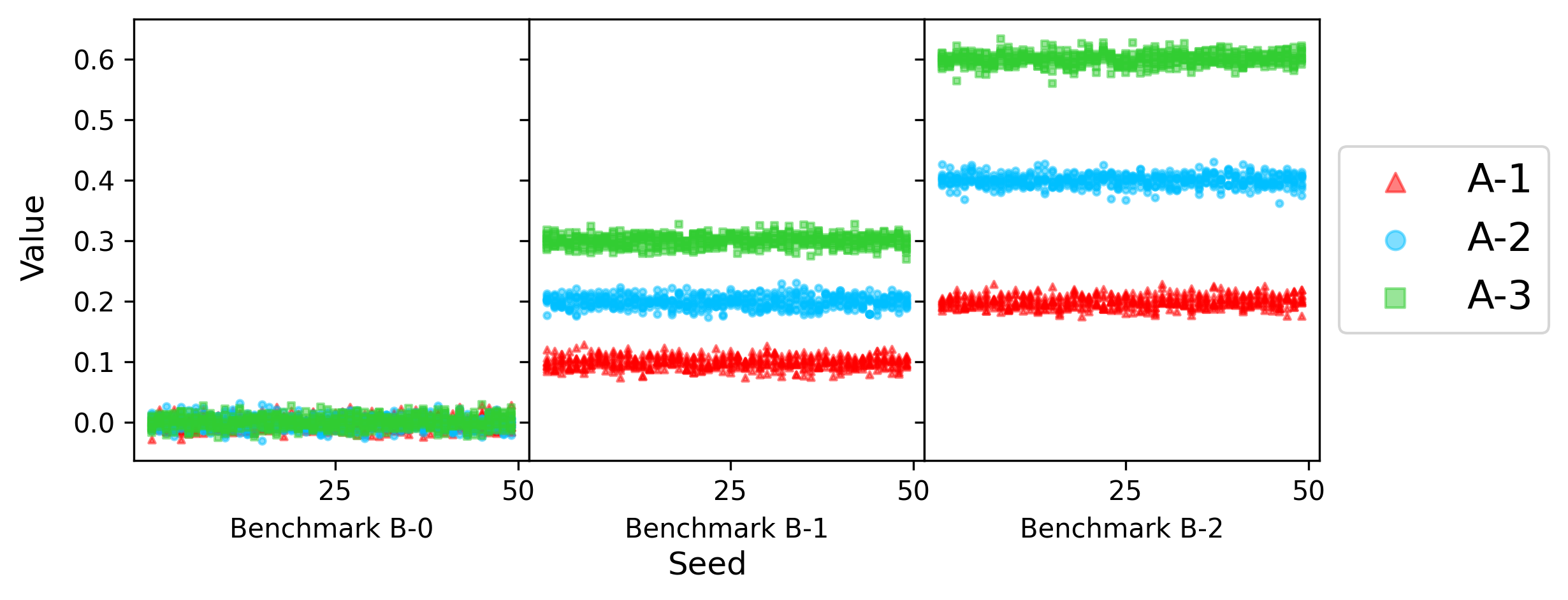}
    \caption[Dataset with varying benchmark-algorithm relations]{\textbf{Multi-Benchmark dataset:} For Benchmark B-0, all algorithms perform similar, for B-1 and B-2, performances vary.}
    \label{fig:benchmark_df}
\end{centering}
\end{figure}

Additionally (but more compute-intensively) we can rank the benchmarks according to their relevance, by using a model with individual algorithm-benchmark random effects (Figure \ref{eq:benchmark_rank_model}) and again looking at the variance of these effects, where higher variance corresponds to larger performance differences.
\begin{equation}\label{eq:benchmark_rank_model}
    \textbf{loss} \boldsymbol{\sim} \textbf{(0 + benchmark|algorithm)}
\end{equation}
\begin{figure}[htpb]
\begin{centering}
    \includegraphics[width=0.7\textwidth]{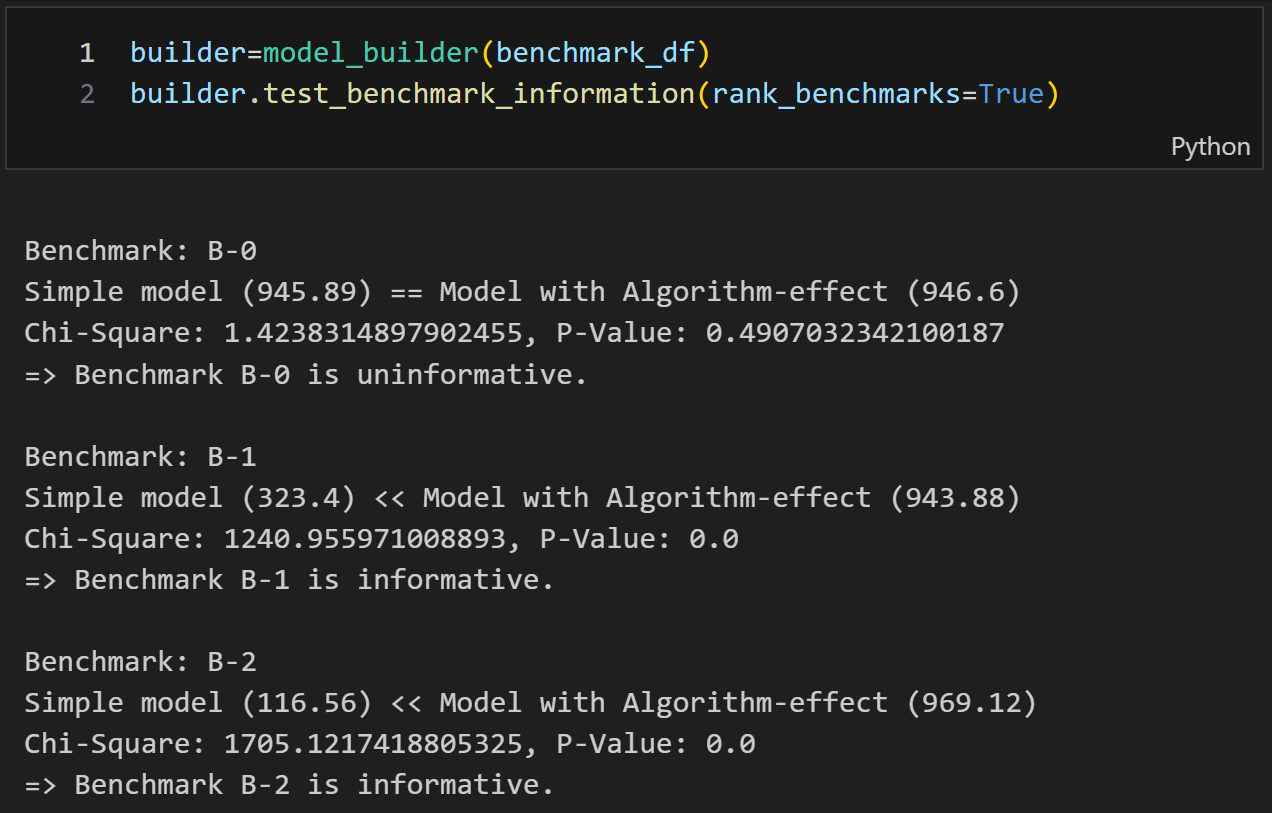}
    \includegraphics[width=0.29\textwidth]{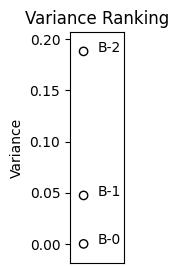}
    \caption[Code: Benchmark Relevance Test]{\textbf{Benchmark relevance test} Left: Individual tests, Right: Variance ranking}
    \label{fig:benchmark_code}
\end{centering}
\end{figure}

As Figure \ref{fig:benchmark_code} shows, benchmark B-0 is uninformative as in the algorithms do not vary significantly within the benchmark and the ranking suggests the largest performance gap for B-2, which corresponds to our data.

\paragraph{Budget relevance}\label{par:budget} LMEMs can natively integrate the HPO budget into the model training process.
\begin{figure}[htpb]
\begin{centering}
    \includegraphics[width=\textwidth]{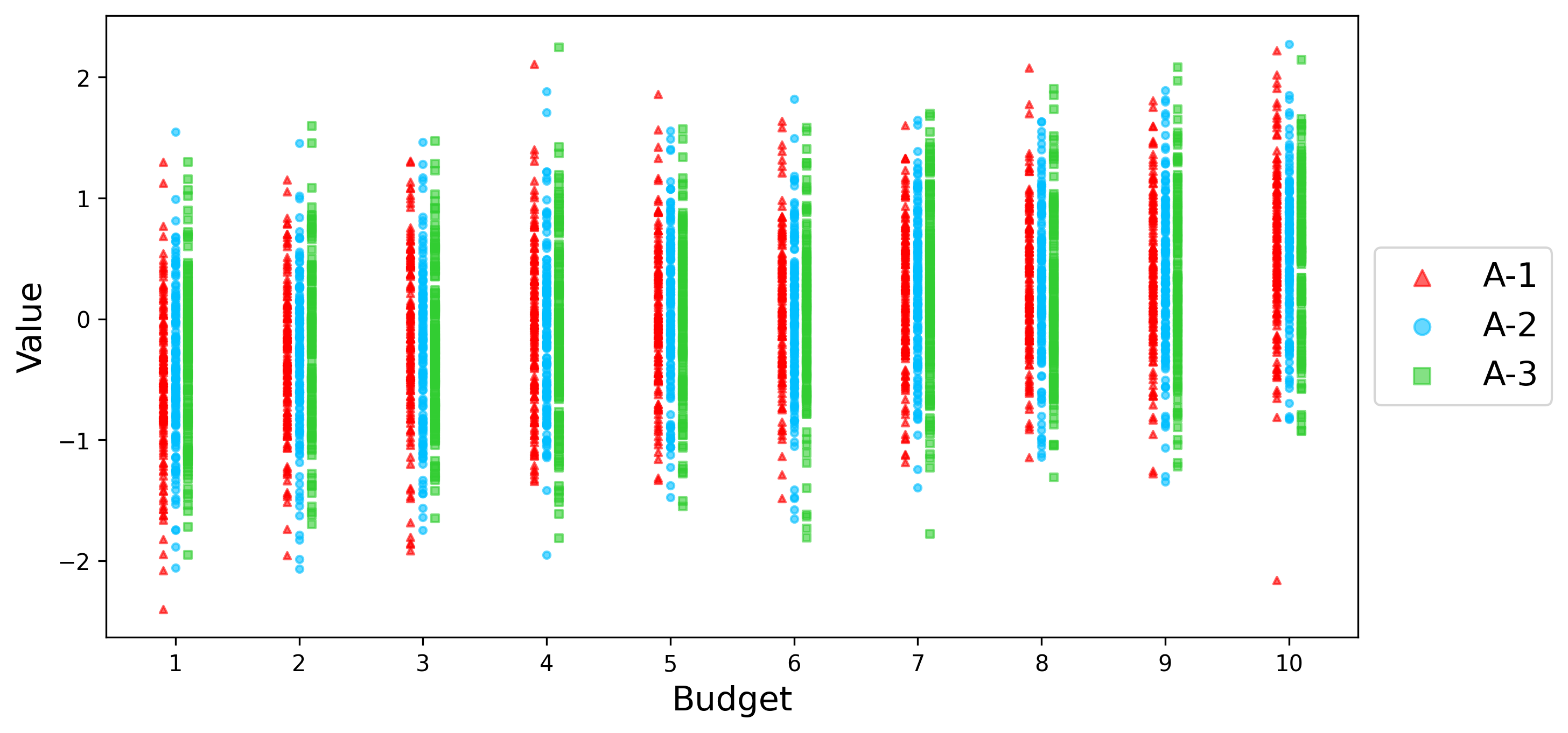}
    \caption[Dataset with Budget-dependent algorithms]{\textbf{Budget dependent dataset:} The training budget loosely affects the mean of each algorithm.}
    \label{fig:fidelity_df}
\end{centering}
\end{figure}
To test whether a fidelity or other Meta-Parameter variation helps understand the data, we propose a set of GLRTs that compares a simple model (Equation \ref{eq:fidelity_simple_model}), a model with fixed fidelity effect (Equation \ref{eq:fidelity_effect_model}) and a model with interaction effect (Equation \ref{eq:fidelity_interaction_model}) to each other. 
\begin{equation}\label{eq:fidelity_simple_model}
    \textbf{loss} \boldsymbol{\sim} \textbf{algorithm} \boldsymbol{+} \textbf{(1|benchmark)}
\end{equation}
\begin{equation}\label{eq:fidelity_effect_model}
    \textbf{loss} \boldsymbol{\sim} \textbf{algorithm} \boldsymbol{+} \textbf{budget} \boldsymbol{+} \textbf{(1|benchmark)}
\end{equation}
\begin{equation}\label{eq:fidelity_interaction_model}
    \textbf{loss} \boldsymbol{\sim} \textbf{algorithm} \boldsymbol{+} \textbf{algorithm:budget} \boldsymbol{+} \textbf{(1|benchmark)}
\end{equation}
\begin{figure}[htpb]
\begin{centering}
    \includegraphics[width=0.8\textwidth]{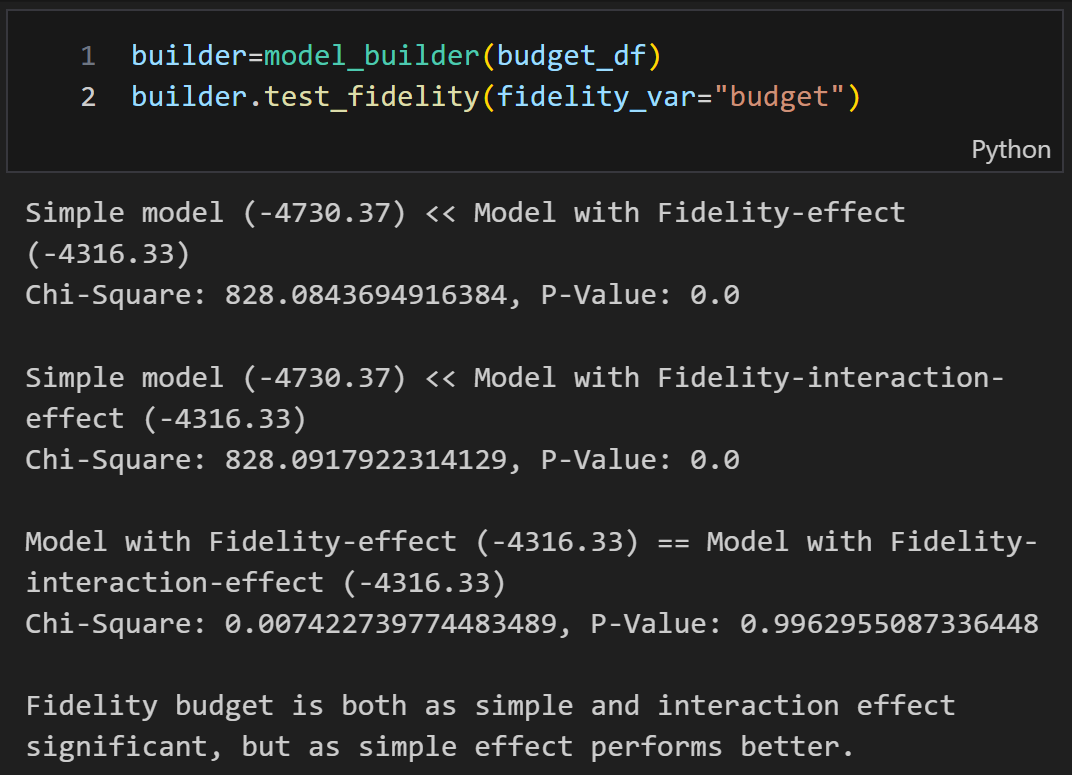}
    \caption[Code: Fidelity Effect Test]{\textbf{Testing budget effect} After comparing models with and without the effects, we conclude that budget is significantly improving the model both as a simple and as an interaction effect. However, as an interaction effect introduces higher complexity, the simple effect performed better of the two.}
    \label{fig:fidelity_code}
\end{centering}
\end{figure}


\end{document}